\pdfoutput=1

\documentclass[11pt]{article}

\usepackage{acl}
\usepackage{times}
\usepackage{latexsym}
\usepackage{mathtools}
\usepackage{amsmath}
\usepackage{txfonts}
\usepackage{xspace}
\usepackage{multirow}
\usepackage{enumitem} 
\newcommand{\macrof}{$\mathrm{m}$-$\mathrm{F_1}$\xspace}
\newcommand{\microf}{$\mathrm{\muup}$-$\mathrm{F_1}$\xspace}

\newcommand{\cls}{\texttt{\small [cls]}\xspace}
\newcommand{\sep}{\texttt{\small [sep]}\xspace}
\newcommand{\specialcell}[2][l]{%
  \begin{tabular}[#1]{@{}l@{}}#2\end{tabular}}

\newcommand{\specialcelll}[2][l]{%
  \begin{tabular}[#1]{@{}p{20cm}@{}}#2\end{tabular}}
\usepackage{microtype}

\title{LexGLUE: A Benchmark Dataset for\\ Legal Language Understanding in English}

\author{Ilias Chalkidis$^{\;\alpha}$\thanks{\;\;Corresponding author: \texttt{ilias.chalkidis@di.ku.dk}}\qquad Abhik Jana$^{\;\beta}$ \qquad Dirk Hartung$^{\;\gamma\;\delta}$ \qquad \textbf{Michael Bommarito$^{\;\gamma\;\delta}$} \\ 
\textbf{Ion Androutsopoulos$^{\;\epsilon}$} \qquad \textbf{Daniel Martin Katz$^{\;\gamma\;\delta\;\zeta}$} \qquad \textbf{Nikolaos Aletras$^{\;\eta}$} \\
$^{\alpha\;}$University of Copenhagen, Denmark \qquad 
$^{\beta\;}$Universität Hamburg, Germany \\
$^\gamma$ Bucerius Law School, Hamburg, Germany \qquad
$^\delta$ CodeX, Stanford Law School, United States \\
$^\epsilon$ Athens University of Economics and Business, Greece \qquad
$^\eta$ University of Sheffield, UK \\
$^\zeta$ Illinois Tech – Chicago Kent College of Law, United States \\
}

\date{}

\begin{document}
\maketitle
\begin{abstract}

Laws and their interpretations, legal arguments and agreements\ are typically expressed in writing, leading to the production of vast corpora of legal text. Their analysis, which is at the center of legal practice, becomes increasingly elaborate as these collections grow in size. Natural language understanding (NLU) technologies can be a valuable tool to support legal practitioners in these endeavors. Their usefulness, however, largely depends on whether current state-of-the-art models can generalize across various tasks in the legal domain. To answer this currently open question, we introduce the Legal General Language Understanding Evaluation (LexGLUE) benchmark, a collection of datasets for evaluating model performance across a diverse set of legal NLU tasks in a standardized way. We also provide an evaluation and analysis of several generic and legal-oriented models demonstrating that the latter consistently offer performance improvements across multiple tasks.
\end{abstract}

\section{Introduction}

Law is a field of human endeavor dominated by the use of language. As part of their professional training, law students consume large bodies of text as they seek to tune their understanding of the law and its application to help manage human behavior. Virtually every modern legal system produces massive volumes of textual data \cite{katz2020}.  Lawyers, judges, and regulators continuously author legal documents such as briefs, memos, statutes, regulations, contracts, patents and judicial decisions \cite{coupette2021}.  Beyond the consumption and production of language, law and the art of lawyering is also an exercise centered around the analysis and interpretation of text. 

Natural language understanding (NLU) technologies can assist legal practitioners in a variety of legal tasks \cite{chalkidis2018deep,aletras2019proceedings, aletras2020proceedings,zhong2020does,bommarito2021Lex}, from judgment prediction \cite{aletras2016predicting,sim-etal-2016-friends,katz2017general,Zhong2018,chalkidis-etal-2019-neural,Malik2021}, information extraction from legal documents \cite{chalkidis-etal-2018-obligation,Chalkidis2019neurips,chen-etal-2020-joint-entity,Hendrycks2021CUAD} and case summarization \cite{Bhattacharya2019} to legal question answering \cite{ravichander-etal-2019-question,kien-etal-2020-answering,Zhong_2020_Iteratively,Zhong2020JECQA} and text classification \cite{nallapati-manning-2008-legal,chalkidis2019Large,chalkidis2020-lmtc}. Transformer models \cite{Vaswani2017} pre-trained on legal, rather than generic, corpora have also been studied \cite{chalkidis-etal-2020-legalbert,zhengguha2021,xiao-etal-2021}.

\begin{figure}[t]
    \centering
    \resizebox{\columnwidth}{!}{
    \includegraphics{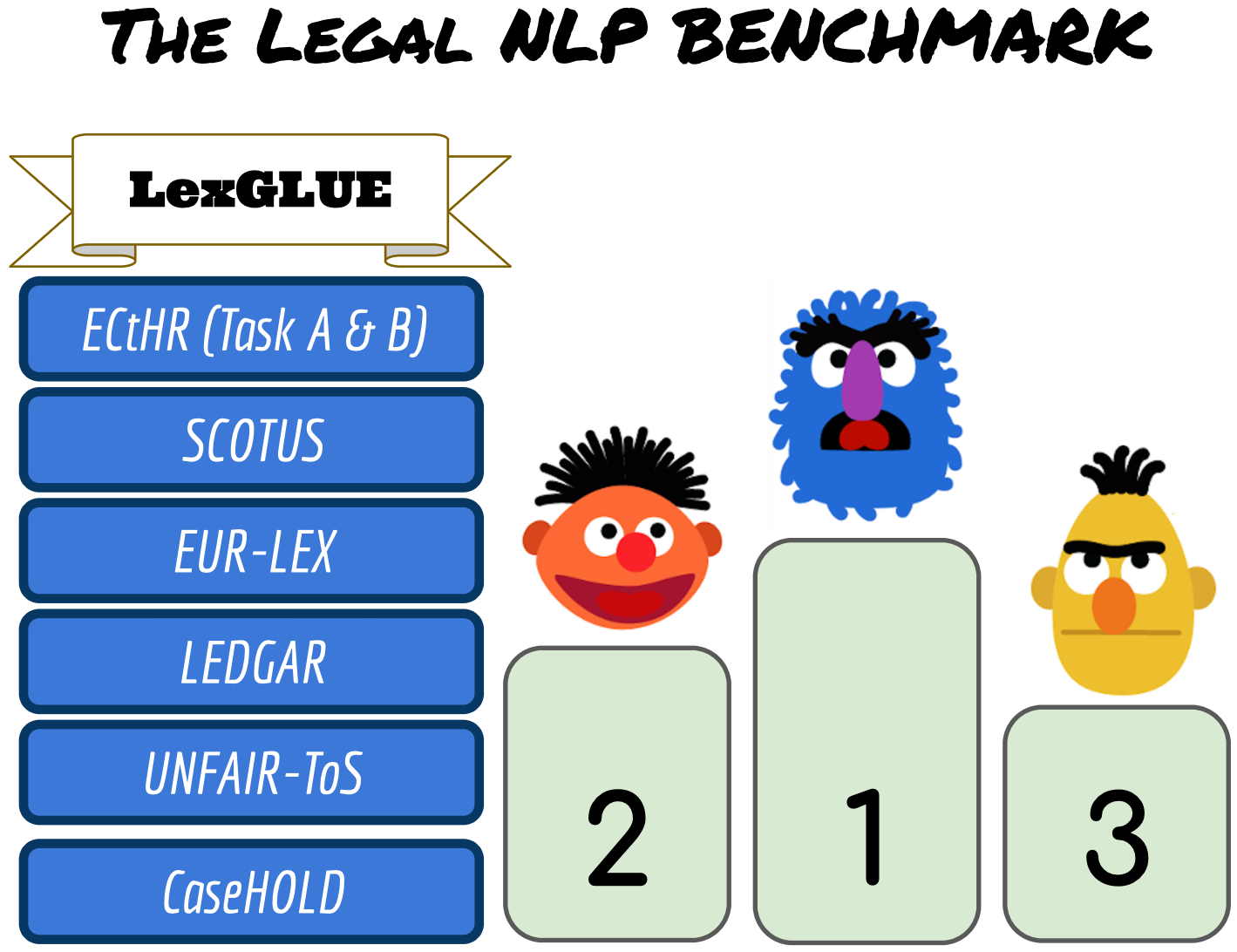}
    }
    \vspace*{-7mm}
    \caption{LexGLUE: A new benchmark dataset to evaluate the capabilities of NLU models on legal text.
    }
    \label{fig:my_label}
    \vspace{-5mm}
\end{figure}

Pre-trained Transformers, including BERT \cite{devlin-etal-2019-bert}, GPT-3 \cite{brown2020language}, T5 \cite{JMLR:v21:20-074}, BART \cite{lewis-etal-2020-bart}, DeBERTa \cite{he2021deberta} and numerous variants, are currently the state of the art in most natural language processing (NLP) tasks. Rapid performance improvements have been witnessed, to the extent that ambitious multi-task benchmarks  \cite{wang-2018-glue,wang-2019-glue} are considered almost `solved' a few years after their release and  need to be made more challenging \cite{wang-2019-superglue}. 

Recently, \citet{bommasani2021opportunities} named these pre-trained models (e.g., BERT, DALL-E, GPT-3) \emph{foundation models}. The term may be controversial, but it emphasizes the paradigm shift these models have caused and their interdisciplinary potential. Studying the latter includes the question of how to adapt these models to legal text \cite{bommarito2021Lex}. As discussed by \citet{zhong2020does} and \citet{chalkidis-etal-2020-legalbert}, legal text has distinct characteristics, such as terms that are uncommon in generic corpora (e.g., ‘restrictive covenant’, ‘promissory estoppel’, ‘tort’, ‘novation’), terms that have different meanings than in everyday language (e.g., an ‘executed’ contract is signed and effective, a `party' is a legal entity), older expressions (e.g., pronominal adverbs like ‘herein’, ‘hereto’, ‘wherefore’), uncommon expressions from other languages (e.g., ‘laches’, ‘voir dire’, ‘certiorari’, ‘sub judice’), and long sentences with unusual word order (e.g., ``the provisions for termination hereinafter appearing or will at the cost of the borrower forthwith comply with the same'') to the extent that legal language is often classified as a `sublanguage' \citep{Tiersma1999,Williams2007,Haigh2018}. Furthermore, legal documents are often much longer than the maximum length state-of-the-art deep learning models can handle, including those designed to handle long text \cite{Longformer,BigBird,yang-etal-2020-smith}. 



Inspired by the recent widespread use of the GLUE multi-task benchmark NLP dataset \cite{wang-2018-glue,wang-2019-glue}, the subsequent more difficult SuperGLUE \cite{wang-2019-superglue}, other previous multi-task NLP benchmarks \cite{conneau-kiela-2018-senteval,DecaNLP}, and similar initiatives in other domains \cite{peng2019transfer}, we introduce LexGLUE, a benchmark dataset to evaluate the performance of NLP methods in legal tasks. LexGLUE is based on seven English existing legal NLP datasets, selected using criteria largely from SuperGLUE (discussed in Section~\ref{sec:desiderata}).

We anticipate that more datasets, tasks, and languages will be added in later versions of LexGLUE.\footnote{See \url{https://nllpw.org/resources/} and \url{https://github.com/thunlp/LegalPapers} for lists of papers, datasets, and other resources related to NLP for legal text.} As more legal NLP datasets become available, we also plan to favor datasets checked thoroughly for validity (scores reflecting real-life performance), annotation quality, statistical power, and social bias \cite{bowman-dahl-2021}.

As in GLUE and SuperGLUE \cite{wang-2019-glue,wang-2019-superglue}, one of our goals is to push towards generic (or `foundation') models that can cope with multiple NLP tasks, in our case legal NLP tasks, possibly with limited task-specific fine-tuning.
Another goal is to provide a convenient and informative entry point for NLP researchers and practitioners wishing to explore or develop methods for legal NLP. Having these goals in mind, the datasets we include in LexGLUE and the tasks they address have been simplified in several ways, discussed below, to make it easier for newcomers and generic models to address all tasks.
We provide Python APIs integrated with Hugging Face \cite{wolf-etal-2020-transformers, lhoest2021datasets} to easily import all the datasets we experiment with and evaluate the performance of different models (Section~\ref{sec:resources}).

By unifying and facilitating the access to a set of law-related datasets and tasks, we hope to attract not only more NLP experts, but also more interdisciplinary researchers (e.g., law doctoral students willing to take NLP courses). 
More broadly, we hope LexGLUE will speed up the adoption and transparent evaluation of new legal NLP methods and approaches in the commercial sector, too.  Indeed, there have been many commercial press releases in the legal tech industry on high-performing systems, but almost no independent evaluation of the performance of machine learning and NLP-based tools. 
A standard publicly available benchmark would also allay concerns of undue influence in predictive models, including the use of metadata which the relevant law expressly disregards. 


\section{Related Work}

The rapid growth of the legal text processing field is demonstrated by numerous papers presented in top-tier conferences in NLP and artificial intelligence~\cite{luo-etal-2017-learning,Zhong2018,chalkidis-etal-2019-neural,valvoda-etal-2021-precedent} as well as surveys~\cite{chalkidis2018deep,zhong2020does,bommarito2021Lex}. Moreover, specialized workshops on NLP for legal text \cite{aletras2019proceedings,di2020ieee,aletras2020proceedings} are regularly organized.

A core task in this area has been legal judgment prediction (forecasting), where the goal is to predict the outcome (verdict) of a court case. In this direction, there have been at least three lines of work. The first one \cite{aletras2016predicting,chalkidis-etal-2019-neural,medvedeva2020using,medvedeva2021automatic} predicts violations of human rights in cases of the European Court of Human Rights (ECtHR). The second line of work \cite{luo-etal-2017-learning, Zhong2018,yang-etal-2019-ljp} considers Chinese criminal cases where the goal is to predict relevant law articles, criminal charges, and the term of the penalty. 
The third line of work \cite{ruger2004supreme, katz2017general,kaufman_kraft_sen_2019} includes methods for predicting the outcomes of cases of the Supreme Court of the United States (SCOTUS). 

The same or similar tasks have also been studied with court cases in many other jurisdictions including France \cite{sulea_predicting_2017}, Philippines \cite{virtucio2018predicting}, Turkey \cite{mumcuouglu2021natural}, Thailand \cite{kowsrihawat2018predicting}, United Kingdom \cite{strickson2020legal}, Germany \cite{urchs_design_2021}, and Switzerland \cite{niklaus-etal-2021-swiss}. Apart from predicting court decisions, there is also work aiming to interpret (explain) the decisions of particular courts \cite{ye_interpretable_2018,chalkidis-et-al-2021-ecthr, branting2021scalable}.

Another popular task is legal topic classification. \citet{nallapati-manning-2008-legal}  highlighted the challenges of legal document classification compared to more generic text classification by using a dataset including docket entries of US court cases. \citet{chalkidis2020-lmtc} classify EU laws into EuroVoc concepts, a task earlier introduced by \citet{Mencia2007}, with a special interest in few- and zero-shot learning. \citet{luz-de-araujo-etal-2020-victor} also studied topic classification using a dataset of Brazilian Supreme Court cases. There are similar interesting applications in contract law~\cite{lippi-etal-2019-claudette,tuggener-etal-2020-ledgar}.

Several studies \cite{chalkidis-etal-2018-obligation,Chalkidis2019neurips,Hendrycks2021CUAD} explored information extraction from contracts, to extract important information such as the contracting parties, agreed payment amount, start and end dates, applicable law, etc. Other studies focus on extracting information from legislation~\cite{cardellino_legal_2017, Angelidis2018NamedER} or court cases~\cite{leitner-2019-ner}.

\begin{table*}[t]
    \centering
    \resizebox{\textwidth}{!}{
    \begin{tabular}{l|l|c|l|c|c}
         \bf Dataset & \bf Source & \bf Sub-domain & \bf Task Type & \bf Training/Dev/Test Instances & \bf  Classes \\
         \hline
         ECtHR (Task A) &\citet{chalkidis-etal-2019-neural} & ECHR & Multi-label classification & 9,000/1,000/1,000 & 10+1\\
         ECtHR (Task B) & \citet{chalkidis-et-al-2021-ecthr}  & ECHR & Multi-label classification & 9,000/1,000/1,000 & 10+1  \\
         SCOTUS & \citet{spaeth2020} & US Law & Multi-class classification & 5,000/1,400/1,400 & 14  \\
         EUR-LEX & \citet{chalkidis2021-multieurlex}  & EU Law & Multi-label classification & 55,000/5,000/5,000 & 100 \\
         LEDGAR & \citet{tuggener-etal-2020-ledgar} & Contracts & Multi-class classification & 60,000/10,000/10,000 & 100 \\
         UNFAIR-ToS & \citet{lippi-etal-2019-claudette} & Contracts & Multi-label classification & 5,532/2,275/1,607 & 8+1 \\
         CaseHOLD & \citet{zhengguha2021} & US Law & Multiple choice QA & 45,000/3,900/3,900 & n/a \\
    \end{tabular}
    }
    \vspace{-2mm}
    \caption{Statistics of the LexGLUE datasets, including simplifications made.}
    \label{tab:examined_datasets}
    \vspace{-5mm}
\end{table*}

Legal Question Answering (QA) is another task of interest in legal NLP, where the goal is to train models for answering legal questions~\cite{Kim2015,ravichander-etal-2019-question,kien-etal-2020-answering,Zhong_2020_Iteratively,Zhong2020JECQA,louis2022statutory}. Not only is this task interesting for researchers but it could support efforts to help laypeople better understand their legal rights. In the general task setting, this requires identifying relevant legislation, case law, or other legal documents, and extracting elements of those documents that answer a particular question. A notable venue for legal QA has been the Competition on Legal Information Extraction and Entailment (COLIEE) \cite{coliee2016,coliee2017,coliee2018}.

More recently, there have also been  efforts to pre-train Transformer-based language models on legal corpora \cite{chalkidis-etal-2020-legalbert,zhengguha2021,xiao-etal-2021}, leading to state-of-the-art results in several legal NLP tasks, compared to models pre-trained on generic corpora.

Overall, the legal NLP literature is overwhelming, and the resources are scattered.
Documentation is often not available, and evaluation measures vary across articles studying the same task. Our goal is to create the first unified benchmark to access the performance of NLP models on legal NLU. As a first step, we selected a representative group of tasks, using datasets in English that are also publicly available, adequately documented and have an appropriate size for developing modern NLP methods. We also introduce several simplifications to make the new benchmark more standardized and easily accessible, as already noted.

\section{LexGLUE Tasks and Datasets}
\label{sec:TasksAndDatasets}

We present the Legal General Language Understanding\footnote{The term `understanding' is, of course, as debatable as in NLU and GLUE, but is commonly used in NLP to refer to systems that analyze, rather than generate text.} Evaluation (LexGLUE) benchmark, a collection of datasets for evaluating model performance across a diverse set of legal NLU tasks.

\subsection{Dataset Desiderata} 
\label{sec:desiderata}
The datasets of LexGLUE were selected to satisfy the following desiderata: 
\begin{itemize}[leftmargin=8pt]
    \item \textbf{Language:} In this first version of LexGLUE, we only consider English datasets, which also makes experimentation easier for researchers across the globe. We hope to include other languages in future versions of LexGLUE. 

    \item \textbf{Substance:}\footnote{We reuse this term from the work of \citet{wang-2019-superglue}.} The datasets should check the ability of systems to understand and reason about 
    legal text to a certain extent in order to perform tasks that are meaningful for legal practitioners. 

    \item \textbf{Difficulty:} The performance of state-of-the-art methods on the datasets should leave large scope for improvements (cf.\ GLUE and SuperGLUE, where top-ranked models now achieve average scores higher than 90\%). Unlike SuperGLUE \cite{wang-2019-superglue}, we did not rule out, but rather favored, datasets requiring domain (in our case legal) expertise. 

    \item \textbf{Availability \& Size:} We consider only publicly available datasets, documented by published articles, avoiding proprietary, untested, poorly documented datasets.  We also excluded very small datasets, e.g., with fewer than 5K documents. Although large pre-trained models often perform well with relatively few task-specific training instances, newcomers may wish to experiment with simpler models that may perform disappointingly with small training sets. Small test sets may also lead to unstable and unreliable results.
\end{itemize}
\vspace{-2mm}

\subsection{Tasks and Datasets}
\label{sec:tasks}
LexGLUE comprises seven datasets. Table~\ref{tab:examined_datasets} shows core information for each of the LexGLUE datasets and tasks, described in detail below.\footnote{In Appendix~\ref{sec:dataset_examples}, we provide examples, i.e., pairs of (inputs, outputs), for all datasets and tasks.}

\paragraph{ECtHR Tasks A \& B} The European Court of Human Rights (ECtHR) hears allegations that a state has breached human rights provisions of the European Convention of Human Rights (ECHR). We use the dataset of \citet{chalkidis-etal-2019-neural, chalkidis-et-al-2021-ecthr}, which contains approx.\ 11K cases from the ECtHR public database. The cases are chronologically split into training (9k, 2001--2016), development (1k, 2016--2017), and test (1k, 2017--2019). For each case, the dataset provides a list of \emph{factual} paragraphs (facts) from the case description. Each case is mapped to \emph{articles} of the ECHR that were violated (if any). In Task A, the input to a model is the list of facts of a case, and the output is the set of violated articles. In the most recent version of the dataset \cite{chalkidis-et-al-2021-ecthr}, each case is also mapped to articles of ECHR that were \emph{allegedly} violated (considered by the court). In Task B, the input is again the list of facts of a case, but the output is the set of allegedly violated articles.

The total number of ECHR articles is currently 66.
Several articles, however,  cannot be violated, are rarely (or never) discussed in practice, or do not depend on the facts of a case and concern procedural technicalities. Thus, we use a simplified version of the label set (ECHR articles) in both Task A and B, including only 10 ECHR articles that can be violated and depend on the case's facts. 

\begin{table*}[t]
    \centering
    \resizebox{\textwidth}{!}{
    \begin{tabular}{l|l|c|c|c|c|ll}
         {\bf Method} & \bf{Source} & \bf \# Params & \bf Vocab. Size & \bf Max Length & \bf Pretrain Specs & \multicolumn{2}{c}{\bf Pre-training Corpora} \\
         \hline
         BERT & \cite{devlin-etal-2019-bert} & 110M & 32K & 512 & 1M / 256 & (16GB) & Wiki, BC \\
         RoBERTa & \cite{liu-2019-roberta} & 125M & 50K & 512 & 100K / 8K & (160GB) & Wiki, BC, CC-News, OWT  \\
         DeBERTa & \cite{he2021deberta} & 139M & 50K & 512 & 1M / 256 & (160GB) & Wiki, BC, CC-News, OWT  \\
         Longformer* & \cite{Longformer} & 149M & 50K & 4096 & 65K / 64 & (160GB) & Wiki, BC, CC-News, OWT \\
         BigBird* & \cite{BigBird} & 127M & 50K & 4096 & 1M / 256 & (160GB) & Wiki, BC, CC-News, OWT \\
         Legal-BERT & \cite{chalkidis-etal-2020-legalbert} & 110M & 32K & 512 & 1M /256 & (12GB) & Legislation, Court Cases, Contracts \\
         CaseLaw-BERT & \cite{zhengguha2021} & 110M & 32K & 512 & 2M /256 & (37GB) & US Court Cases \\
    \end{tabular}
    }
    \vspace{-3mm}
    \caption{Key specifications of the examined models. We report the number of parameters, the size of vocabulary, the maximum sequence length, the core pre-training specifications (training steps and batch size), and the training corpora (OWT = OpenWebText, BC = BookCorpus). Starred models have been warm-started from RoBERTa.}
    \label{tab:examined_models}
    \vspace{-5mm}
\end{table*}

\paragraph{SCOTUS}

The US Supreme Court  (SCOTUS)\footnote{\url{https://www.supremecourt.gov}} is the highest federal court in the United States of America and generally hears only the most controversial or otherwise complex cases which have not been sufficiently well solved by lower courts. We release a new dataset combining information from SCOTUS opinions\footnote{\url{https://www.courtlistener.com}} with the Supreme Court DataBase (SCDB)\footnote{\url{http://scdb.wustl.edu}} \cite{spaeth2020}. SCDB provides metadata (e.g., decisions, issues, decision directions) for all cases (from 1946 up to 2020).
We opted to use SCDB to classify the court  \emph{opinions} in the available 14  \emph{issue areas} (e.g., Criminal Procedure, Civil Rights, Economic Activity, etc.). This is a single-label multi-class classification task (Table~\ref{tab:examined_datasets}). The 14 issue areas cluster 278 issues whose focus is on the subject matter of the controversy (dispute). The SCOTUS cases are chronologically split into training (5k, 1946--1982), development (1.4k, 1982--1991), test (1.4k, 1991--2016) sets.
\vspace{-1mm}

\paragraph{EUR-LEX} European Union (EU) legislation is published in the EUR-Lex portal.\footnote{\url{http://eur-lex.europa.eu/}} All EU laws are annotated by EU's Publications Office with multiple concepts from EuroVoc, a multilingual thesaurus maintained by the Publications Office.\footnote{\url{http://eurovoc.europa.eu/}} 
The current version of EuroVoc contains more than 7k concepts referring to various activities of the EU and its Member States (e.g., economics, health-care, trade). 
We use the English part of the dataset of \citet{chalkidis2021-multieurlex}, which comprises 65k EU laws (documents) from EUR-Lex. Given a document, the task is to predict its EuroVoc labels (concepts).
The dataset is chronologically split in training (55k, 1958--2010), development (5k, 2010--2012), test (5k, 2012--2016) subsets. It supports four different label granularities,  comprising 21, 127, 567, 7390 EuroVoc concepts, respectively. 
We use the 100 most frequent concepts from level 2, which has a highly skewed label distribution and temporal concept drift \cite{chalkidis2021-multieurlex}, making it sufficiently difficult for an entry point.   

\paragraph{LEDGAR}

\citet{tuggener-etal-2020-ledgar} introduced LEDGAR (Labeled EDGAR), a dataset for contract provision (paragraph) classification. The contract provisions come from contracts obtained from the US Securities and Exchange Commission (SEC) filings, which are publicly available from EDGAR\footnote{\url{https://www.sec.gov/edgar/}} (Electronic Data Gathering, Analysis, and Retrieval system).
The original dataset includes approx.\ 850k contract provisions labeled with 12.5k categories. Each label represents the single main topic (theme) of the corresponding contract provision, i.e., this is a single-label multi-class classification task. In LexGLUE, we use a subset of the original dataset with 80k contract provisions, considering only the 100 most frequent categories as a simplification. We split the new dataset chronologically into training (60k, 2016--2017), development (10k, 2018), and test (10k, 2019) sets.

\paragraph{UNFAIR-ToS} The UNFAIR-ToS dataset \cite{lippi-etal-2019-claudette} contains 50 Terms of Service (ToS) from on-line platforms (e.g., YouTube, Ebay, Facebook, etc.). The dataset has been annotated on the sentence-level with 8 types of \emph{unfair contractual terms}, meaning terms (sentences) that potentially violate user rights according to EU consumer law.\footnote{Art.\ 3 of Direct.\ 93/13, Unfair Terms in Consumer Contracts (\url{http://data.europa.eu/eli/dir/1993/13/oj}).} The input to a model is a sentence, the output is the set of unfair types (if any). We split the dataset chronologically into training (5.5k, 2006--2016), development (2.3k, 2017),  test (1.6k, 2017) sets.

\paragraph{CaseHOLD}
The CaseHOLD (Case Holdings on Legal Decisions) dataset \cite{zhengguha2021} contains approx. 53k multiple choice questions about holdings of US court cases from the Harvard Law Library case law corpus. \emph{Holdings} are short summaries of legal rulings that accompany referenced decisions relevant for the present case, e.g.:\vspace{1mm}

\noindent
\emph{``$\dots$ to act pursuant to City policy, re d 503, 506-07 (3d Cir.l985)(\textbf{holding that for purposes of a class certification motion the court must accept as true all factual allegations in the complaint and may draw reasonable inferences therefrom}).''}\vspace{1mm}

The input consists of an \emph{excerpt} (or prompt) \emph{from a court decision}, containing a reference to a particular case, where the \emph{holding} statement (in boldface) is masked out.
The model must identify the correct (masked) holding statement from a selection of five choices.
We split the dataset in training (45k), development (3.9k), test (3.9k) sets, excluding samples that are shorter than 256 tokens. Chronological information is missing from CaseHOLD, thus we cannot perform a chronological re-split.

\begin{figure*}
    \centering
    \resizebox{\textwidth}{!}{
    \includegraphics{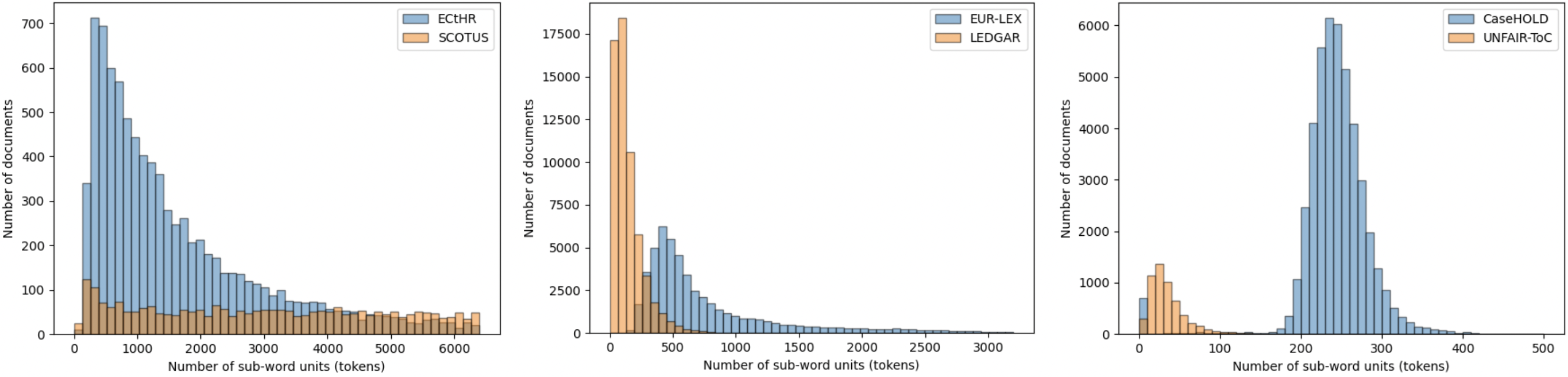}
    }
    \vspace{-6mm}
    \caption{Distribution of text input length, measured in BERT sub-word units, across LexGLUE datasets.}
    \label{fig:lengths}
    \vspace{-5mm}
\end{figure*}

\section{Models Considered}

\subsection{Linear SVM}
Our first baseline model is a linear Support Vector Machine (SVM) \cite{cortes-1995} with TF-IDF features for the top-$K$ frequent $n$-grams of the training set, where $n\in[1,2,3]$.

\subsection{Pre-trained Transformer Models} \label{sec:pretrainedModels}
\label{sec:baselines}
We experiment with Transformer-based \cite{Vaswani2017} pre-trained language models, which achieve state of the art performance in most NLP tasks \cite{bommasani2021opportunities} and NLU benchmarks~\cite{wang-2019-superglue}. These models are pre-trained on very large unlabeled corpora to predict masked tokens (masked language modeling) and typically also to perform other pre-training tasks that still do not require any manual annotation (e.g., predicting if two sentences were adjacent in the corpus or not, dubbed next sentence prediction). The pre-trained models are then fine-tuned (further trained) on task-specific (typically much smaller) annotated datasets, after adding task-specific layers. 
We fine-tune and evaluate the performance of the following publicly available models (Table~\ref{tab:examined_models}).\vspace{2mm}

\noindent\textbf{BERT} \cite{devlin-etal-2019-bert} is the best-known pre-trained Transformer-based language model. It is pre-trained to perform masked language modeling and  next sentence prediction.\vspace{2mm}

\noindent\textbf{RoBERTa} \cite{liu-2019-roberta} is also a pre-trained Transformer-based language model. Unlike BERT, RoBERTa uses dynamic masking, it eliminates the next sentence prediction pre-training task, uses a larger vocabulary, and has been pre-trained on much larger corpora. \citet{liu-2019-roberta} reported improved results on NLU benchmarks using RoBERTa, compared to BERT.\vspace{2mm}

\noindent\textbf{DeBERTa} \cite{he2021deberta} is another improved BERT model that uses disentangled attention, i.e., four separate attention mechanisms considering the content and the relative position of each token, and an enhanced mask decoder, which explicitly considers the absolute position of the tokens. DeBERTa has been reported to outperform BERT and RoBERTa in several NLP tasks \cite{he2021deberta}.\vspace{2mm}

\noindent\textbf{Longformer} \cite{Longformer} extends Transformer-based models to support longer sequences, using sparse-attention. The latter is a combination of local (window-based) attention and global (dilated) attention that reduces the computational complexity of the model and thus can be deployed in longer documents (up to 4096 tokens). Longformer outperforms RoBERTa on long document tasks and QA benchmarks.\vspace{2mm}

\noindent\textbf{BigBird} \cite{BigBird} is another sparse-attention based transformer that uses a combination of a local (window-based) attention, global (dilated), and random attention, i.e., all tokens also attend a number of random tokens on top of those in the same neighborhood (window) and the global ones. BigBird has been reported to outperform Longformer on QA and summarization tasks.\vspace{2mm}

\noindent\textbf{Legal-BERT} \cite{chalkidis-etal-2020-legalbert} is a BERT model pre-trained on English legal corpora, consisting of legislation, contracts, and court cases. It uses the original pre-training BERT configuration. The sub-word vocabulary of Legal-BERT is built from scratch, to better support legal terminology.\vspace{2mm}

\noindent\textbf{CaseLaw-BERT} \cite{zhengguha2021} is another law-specific BERT model. It also uses the original pre-training BERT configuration and has been pre-trained from scratch on the Harvard Law case corpus,\footnote{\url{https://case.law/}} which comprises 3.4M legal decisions from US federal and state courts. This model is called \emph{Custom Legal-BERT} by  \citet{zhengguha2021}. We call it CaseLaw-BERT to distinguish it from the previously published Legal-BERT of \citet{chalkidis-etal-2020-legalbert} and to better signal that it is trained exclusively on case law (court opinions).\vspace{2mm}

\noindent\textbf{Hierarchical Variants} Legal documents are usually much longer (i.e., consisting of thousands of words) than other text types (e.g., tweets, customer reviews, news articles) often considered in various NLP tasks. Thus, standard Transformer-based models that can typically process up to 512 sub-word units cannot be directly applied across all LexGLUE datasets, unless documents are severely truncated to the model's limit.  Figure~\ref{fig:lengths} shows  the distribution of text input length across all LexGLUE datasets. Even for Transformer-based models specifically designed to handle long text (e.g., Longformer, BigBird), handling longer legal documents remains a challenge. 

Given the length of the text input in three of the seven LexGLUE tasks, i.e., ECtHR (A and B) and SCOTUS, we employ a hierarchical variant of each pre-trained Transformer-based model that has not been designed for longer text (BERT, RoBERTa, DeBERTa, Legal-BERT, CaseLaw-BERT) during fine-tuning and inference. The hierarchical variants are similar to those of \citet{chalkidis-et-al-2021-ecthr}. They use the corresponding pre-trained Transformer-based model to encode each paragraph of the input text independently and obtain the top-level representation $h_{\text{\cls}}$ of each paragraph. A second-level shallow (2-layered) Transformer encoder with always the same (across BERT, RoBERTa, DeBERTa etc.) specifications (e.g., hidden units, number of attention heads) is fed with the paragraph representations to make them context-aware (aware of the surrounding paragraphs). We then max-pool over the context-aware paragraph representations to obtain a document representation, which is fed to a classification layer.\footnote{In Appendix~\ref{sec:bert_truncated}, we present results from preliminary experiments using the standard version of BERT for ECtHR Task A (-12.2\%), Task B(-10.6\%), and SCOTUS (-3.5\%).}

\subsection{Task-Specific Fine-Tuning}

\noindent\textbf{Text Classification Tasks} For EUR-LEX, LEDGAR and UNFAIR-ToS tasks, we feed each document to the pre-trained model (e.g., BERT) and obtain the top-level representation $h_{\text{\cls}}$ of the special \cls token as the document representation, following \citet{devlin-etal-2019-bert}. The latter goes through a dense layer of $L$ output units, one per label, followed by a sigmoid (in EUR-LEX, UNFAIR-ToS) or softmax (in LEDGAR) activation, respectively. For the two ECtHR tasks (A and B) and SCOTUS, where the hierarchical variants are employed, we feed the max-pooled (over paragraphs) document representation to a classification linear layer. The linear layer is again followed by a sigmoid (EctHR) or softmax (SCOTUS) activation.\vspace{2mm}

\noindent\textbf{Multiple-Choice QA Task} For CaseHOLD, we convert each training (or test) instance (the prompt and the five candidate answers) into five input pairs following \citet{zhengguha2021}. Each pair consists of the prompt and one of the five candidate answers, separated by the special delimiter token \sep. The top-level representation $h_{\text{\cls}}$ of each pair is fed to a linear layer to obtain a logit, and the five logits are then passed through a softmax yielding a probability distribution over the five candidate answers.

\subsection{Data Repository and Code}
\label{sec:resources}

For reproducibility purposes and to facilitate future experimentation with other models, we pre-process and release all datasets on Hugging Face Datasets \cite{lhoest2021datasets}.\footnote{\url{https://huggingface.co/datasets/lex_glue}}  We also release the code\footnote{\url{https://github.com/coastalcph/lex-glue}} of our experiments, which relies on the Hugging Face Transformers \cite{wolf-etal-2020-transformers} library.\footnote{\url{https://huggingface.co/transformers}} Appendix~\ref{sec:codebase} explains how to load the datasets and run experiments with our code. 

\renewcommand{\arraystretch}{1.2}
\begin{table*}[t]
    \centering
    \resizebox{\textwidth}{!}{
    \begin{tabular}{l|cc|cc|cc|cc|cc|cc|c|c}
         \multirow{2}{*}{\bf Method}  & \multicolumn{2}{c|}{\bf ECtHR (A)*} & \multicolumn{2}{c|}{\bf ECtHR (B)*} & \multicolumn{2}{c|}{\bf SCOTUS*} & \multicolumn{2}{c|}{\bf EUR-LEX} & \multicolumn{2}{c|}{\bf LEDGAR}  & \multicolumn{2}{c|}{\bf UNFAIR-ToS} & \bf CaseHOLD\\
         & \microf & \macrof & \microf & \macrof & \microf & \macrof & \microf & \macrof & \microf & \macrof & \microf & \macrof & \microf / \macrof \\
         \hline
TFIDF-SVM       & 62.6 & 48.9 & 73.0 & 63.8 & 74.0 & 64.4 & 63.4 & 47.9 & 87.0 & 81.4 & 94.7 & 75.0 & 22.4 \\
\hline
BERT            & \bf 71.2 & 63.6 & 79.7 & 73.4 & 68.3 & 58.3 & 71.4 & 57.2 & 87.6 & 81.8 & 95.6 & 81.3 & 70.8 \\
RoBERTa         & 69.2 & 59.0 & 77.3 & 68.9 & 71.6 & 62.0 & 71.9 & \bf 57.9 & 87.9 & 82.3 & 95.2 & 79.2 & 71.4 \\
DeBERTa         & 70.0 & 60.8 & 78.8 & 71.0 & 71.1 & 62.7 & \bf 72.1 & 57.4 & 88.2 & \bf 83.1 & 95.5 & 80.3 & 72.6 \\
\hline
Longformer      & 69.9 & \bf 64.7 & 79.4 & 71.7 & 72.9 & 64.0 & 71.6 & 57.7 & 88.2 & 83.0 & 95.5 & 80.9 & 71.9 \\
BigBird         & 70.0 & 62.9 & 78.8 & 70.9 & 72.8 & 62.0 & 71.5 & 56.8 & 87.8 & 82.6 & 95.7 & 81.3 & 70.8 \\
\hline
Legal-BERT      & 70.0 & 64.0 & \bf 80.4 & \bf 74.7 & 76.4 & \bf 66.5 & \bf 72.1 & 57.4 & 88.2 & 83.0 & \bf 96.0 & \bf 83.0 & 75.3 \\
CaseLaw-BERT    & 69.8 & 62.9 & 78.8 & 70.3 & \bf 76.6 & 65.9 & 70.7 & 56.6 & \bf 88.3 & 83.0 & \bf 96.0 & 82.3 & \bf 75.4 \\
    \end{tabular}
    }
    \vspace{-3mm}
    \caption{Test results for all examined models across LexGLUE tasks. In starred datasets, we use the hierarchical variant of each model, except for Longformer and BigBird, discussed in Section~\ref{sec:baselines}.
    }
    \label{tab:leaderboard}
    \vspace{-4mm}
\end{table*}

\section{Experiments} 

\subsection{Experimental Set Up}

For TFIDF-based linear SVM models, we use the implementation of Scikit-learn \cite{scikit-learn} and grid-search for hyper parameters (number of features, $C$, and loss function).
For all the pre-trained models, we use publicly available Hugging Face checkpoints.\footnote{\url{http://huggingface.co/models}} We use the *-base configuration of each pre-trained model, i.e., 12 Transformer blocks, 768 hidden units, and 12 attention heads. We train models with the Adam optimizer \cite{Kingma2015} and an initial learning rate of 3e-5 up to 20 epochs using early stopping on development data. We use mixed precision (fp16) to decrease the memory footprint in training and gradient accumulation for all hierarchical models.
The hierarchical models can read up to 64 paragraphs of 128 tokens each. We use Longformer and BigBird in default settings, i.e., Longformer uses windows of 512 tokens and a single global token (\cls), while BigBird uses blocks of 64 tokens (windows: 3$\times$ block, random: 3$\times$ block, global: 2$\times$ initial block; each token attends 512 tokens in total).  The batch size is 8 in all experiments. We run five repetitions with different random seeds and report the test scores based on the seed with the best scores on development data. We evaluate performance using \emph{micro-F1} (\microf) and \emph{macro-F1} (\macrof) across all datasets to take into account class imbalance. For completeness, we also report the arithmetic, harmonic, and geometric mean across tasks following \citet{shavrina2021how}.\footnote{We acknowledge that the use of scores aggregated over tasks has been criticized in general NLU benchmarks (e.g., GLUE), as models are trained with different numbers of samples, task complexity, and evaluation metrics per task. We believe that the use of a standard common metric (F1) across tasks and averaging with harmonic mean alleviate this issue.}
\vspace{-1mm}

\begin{table}[t]
    \centering
        \resizebox{\columnwidth}{!}{
    \begin{tabular}{l|cc|cc|cc|}
\multirow{2}{*}{\bf Method}  & \multicolumn{2}{c|}{\bf A-Mean} &  \multicolumn{2}{c|}{\bf H-Mean} &  \multicolumn{2}{c|}{\bf G-Mean} \\
& \microf & \macrof & \microf & \macrof & \microf & \macrof \\
\hline
                               BERT &  77.8 &  69.5 &  76.7 &  68.2 &  77.2 &  68.8 \\
                            RoBERTa &  77.8 &  68.7 &  76.8 &  67.5 &  77.3 &  68.1 \\
                            DeBERTa &  78.3 &  69.7 &  77.4 &  68.5 &  77.8 &  69.1 \\
                            \hline
                         Longformer &  78.5 &  70.5 &  77.5 &  69.5 &  78.0 &  70.0 \\
                            BigBird &  78.2 &  69.6 &  77.2 &  68.5 &  77.7 &  69.0 \\
                            \hline
                         Legal-BERT &  \bf 79.8 &  \bf 72.0 &  \bf 78.9 &  \bf 70.8 &  \bf 79.3 &  \bf 71.4 \\
                       CaseLaw-BERT &  79.4 &  70.9 &  78.5 &  69.7 &  78.9 &  70.3 \\
    \end{tabular}
    }
    \vspace*{-3mm}
    \caption{Test scores aggregated over tasks: arithmetic (A), harmonic (H), and geometric (G) mean.}
    \label{tab:avg_leaderboard}
    \vspace{-5mm}
\end{table}

\subsection{Experimental Results}

\paragraph{Main Results}
Table~\ref{tab:leaderboard} presents the test results for all models across all LexGLUE tasks, while Table~\ref{tab:avg_leaderboard} presents the aggregated (averaged) results. We observe that the two legal-oriented pre-trained models (Legal-BERT, CaseLaw-BERT) perform overall better, especially considering \macrof that accounts for class imbalance (considers all classes equally important). Their in-domain (legal) knowledge seems to be more critical in the two datasets relying on US case law data (SCOTUS, CaseHOLD) with an improvement of approx.\ +2-4\% p.p. (\macrof) over equally sized Transformer-based models, which are pre-trained on generic corpora. These results are explained by the fact that these tasks are more domain-specific in terms of language, compared to the rest.  No single model performs best in all tasks, and the results of Table~\ref{tab:leaderboard} show that there is still large scope for improvement (Section~\ref{sec:vision}).

An exceptional case of the dominance of the pre-trained Transformer models is the SCOTUS dataset, where the TFIDF-based linear SVM performs better than all generic Transformer models. TFIDF-SVM models are domain-specific, since the vocabulary (n-grams) and their IDF scores used to compute TF-IDF scores, are customized per task; which seems to be important for SCOTUS.

\paragraph{Legal-oriented Models}
Interestingly, the performance of Legal-BERT and  CaseLaw-BERT, the two legal-oriented pre-trained models, is almost identical on CaseHOLD, despite the fact that CaseLaw-BERT is solely trained on US case law. On the other hand, Legal-BERT has been exposed to a wider variety of legal corpora, including EU and UK legislation, ECtHR, ECJ and US court cases, and US contracts. Legal-BERT performs as well as or better than CaseLaw-BERT on all datasets. These results suggest that domain-specific pre-training (and learning a domain-specific sub-word vocabulary) is beneficial, but over-fitting a specific (niche) sub-domain (e.g., US case law), similarly to \citet{zhengguha2021}, has no benefits.

\section{Vision -- Future Considerations}
\label{sec:vision}
Beyond the scope of this work and the examined baseline models, we identify four major factors that could potentially advance the state of the art in LexGLUE and legal NLP more generally:\vspace{2mm}

\noindent\textbf{Long Documents:} Several Transformer-based  models~\cite{Longformer, BigBird, liu2022erniesparse} have been proposed to handle long documents by exploring sparse attention mechanisms. These models can handle sequences up to 4096 sub-words, which is largely exceeded in three out of seven LexGLUE tasks (Figure~\ref{fig:lengths}). Contrary, the hierarchical model of Section~\ref{sec:baselines} can handle sequences up to 8192 sub-words in our experiments, but a part of the model (the additional Transformer blocks that make the paragraph embeddings aware of the other paragraphs) is not pre-trained, which possibly negatively affects performance.\vspace{2mm} 

\noindent\textbf{Structured Text:} 
Current models for long documents, like Longformer and BigBird, do not consider the document structure (e.g., sentences, paragraphs, sections). For example, window-based attention may consider a sequence of sentences across paragraph boundaries or even consider truncated sentences.
To exploit the document structure, \citet{yang-etal-2020-smith} proposed SMITH, a hierarchical Transformer model that hierarchically encodes increasingly larger blocks (e.g., words, sentences, documents). SMITH is very similar to the hierarchical model of Section~\ref{sec:baselines}, but it is pre-trained end-to-end with two objectives: token-level masked and sentence block language modeling. \vspace{2mm}

\noindent\textbf{Large-scale Legal Pre-training:} 
Recent studies~\cite{chalkidis-etal-2020-legalbert,zhengguha2021,bambroo-etal-2021,xiao-etal-2021} introduced  language models pre-trained on legal corpora, but of relatively small sizes, i.e., 12--36 GB. In the work of \citet{zhengguha2021}, the pre-training corpus covered only a narrowly defined area of legal documents, US court opinions. The same applies to Lawformer \cite{xiao-etal-2021}, which was pre-trained on Chinese court opinions. 
Future work could curate and release a legal version of the C4 corpus~\cite{JMLR:v21:20-074}, containing multi-jurisdictional legislation, court decisions, contracts and legal literature at a size of hundreds of GBs. Given such a corpus, a large language model capable of processing long structured text could be pre-trained and it might excel in LexGLUE.\vspace{2mm}

\noindent\textbf{Even Larger Language Models:} Scaling up the capacity of pre-trained models has led to increasingly better results in general NLU benchmarks \cite{kaplan2020}, and models have been scaled up to billions of parameters~\cite{brown2020language,JMLR:v21:20-074, he2021deberta}. In Appendix~\ref{sec:roberta-large}, we observe that using the large version of RoBERTa leads to substantial performance improvements compared to the base version. The results are comparable or better - in some cases- compared to the legal-oriented language models (Legal-BERT, CaseLaw-BERT). Considering that the two legal-oriented models are much smaller and have been pre-trained with ($5\!-\!10\times$) less data (Section~\ref{tab:examined_models}), we have a strong indication for performance gains by pre-training larger legal-oriented models using larger legal corpora.

\section{Limitations and Future Work}
\label{sec:limitations}
Although, our benchmark inevitably cannot cover ``\emph{everything in the whole wide (legal) world}'' \cite{raji2021ai}, we include a representative collection of English datasets that also ground to a certain degree in practically interesting applications.

In its current version, LexGLUE can only be used to evaluate English models.
As legal documents are typically written in the official language of the particular country of origin, there is an increasing need for developing models for other languages.
The current scarcity of datasets in other languages (with the exception of Chinese) makes a multilingual extension of LexGLUE challenging, but an interesting avenue for future research. 

Beyond language barriers, legal restrictions currently inhibit the creation of more datasets. 
Important document types, such as contracts and scholarly publications are protected by copyright or considered trade secrets.
As a result, their owners are concerned with data-leakage when they are used for model training and evaluation.
Providing both legal and technical solutions, e.g., using privacy-aware infrastructure and models \cite{downie-2004-imirsel,feyisetan2020privacy} is a challenge to be addressed. 

Access to court decisions can also be hindered by bureaucratic inertia, outdated technology and data protection concerns, which collectively result in these otherwise public decisions not being publicly available \cite{pah2020build}. While the anonymization of personal data provides a solution to this problem, it is itself an open challenge for legal NLP \cite{jana2021investigation}. 
In lack of suitable datasets and benchmarks, we have refrained from including anonymization in this version of LexGLUE, but plan to do so at a later stage.

Another limitation of the current version of LexGLUE is that human evaluation is missing. All datasets rely on \emph{ground truth} labels automatically extracted from data (e.g., court decisions) produced as part of official judicial or archival procedures. These resources should be highly reliable (valid), but we cannot statistically assess their quality. In the future, re-annotating part of the datasets with multiple legal experts would provide an estimation of human level performance and inter-annotator agreement, though the cost would be high, because of the required legal expertise. 

While LexGLUE offers a much needed unified testbed for legal NLU, there are several other critical aspects that need to be studied carefully. These include multi-disciplinary research to better understand the limitations and challenges of applying NLP to law \cite{Binns_2020}, while also considering fairness and robustness \cite{angwin2016,dressel2018,baker-gillis-2021-sexism,wang-etal-2021-equality,chalkidis-2022-fairlex}, and broader legal considerations of AI technologies in general \cite{schwemer2021,tsarapatsanis-aletras-2021-ethical,Delacroix_2022}.

\section*{Acknowledgments}
This work was partly funded by the Innovation Fund Denmark (IFD)\footnote{\url{https://innovationsfonden.dk/en}} under File No.\ 0175-00011A and by the German Federal Ministry of Education and Research (BMBF) kmu-innovativ program under funding code 01IS18085.
We would like to thank Desmond Elliott for providing valuable feedback (baselines for truncated documents presented in Appendix~\ref{sec:bert_truncated}), Xiang Dai and Joel Niklaus for reviewing and pointing out issues in the new resources (code, datasets). 

\section*{Ethics Statement}

\noindent\textbf{Original Work Attribution}\vspace{1mm}

\noindent All datasets included in LexGLUE, except SCOTUS, are publicly available and have been previously published. If datasets or the papers that introduced them were not compiled or written by ourselves, we referenced the original work and encourage LexGLUE users to do so as well. In fact, we believe this work should only be referenced, in addition to citing the original work, when experimenting with multiple LexGLUE datasets and using the LexGLUE evaluation infrastructure. Otherwise only the original work should be cited.\vspace{2mm}

\noindent\textbf{Social Impact}\vspace{1mm}

\noindent We believe that this work does not contain any grounds for ethical concerns. A transparent and rigorous benchmark for NLP in the legal domain might serve as an orientation for scholars and industry researchers. As a result, the capabilities of tools that are trained using natural language data from the legal domain will become clearer, thereby helping their users to better understand them. This increased certainty would also raise the awareness within research and industry communities to potential risks associated with the use of these tools. We regard this contribution to a more realistic, more informed discussion as an important use case of the work presented. Ideally, it could help both beginners and seasoned professionals to understand the limitations of using NLP tools in the legal domain and thereby prevent exaggerated expectations and potential applications that might risk endangering fundamental rights or the rule of law. 
We currently cannot imagine use cases of this particular work that would lead to ethical concerns or potential harm~\cite{tsarapatsanis-aletras-2021-ethical}.\vspace{2mm}

\noindent\textbf{Licensing \& Personal Information}\vspace{1mm}

\noindent LexGLUE comprises seven datasets: ECtHR Task A and B, SCOTUS, EUR-LEX, LEDGAR, UNFAIR-ToS, and CaseHOLD that are available for re-use and re-share with  appropriate attribution. The data is in general partially anonymized in accordance with the applicable national law. The data is considered to be in the public sphere from a privacy perspective. This is a very sensitive matter, as the courts try to keep a balance between transparency (the public's right to know) and privacy (respect for private and family life).

ECtHR contains personal data of the parties and other people involved in the legal proceedings. Its data is processed and made public in accordance with the European data protection laws. This includes either implied consent or legitimate interest to process the data for research purposes. As a result, their processing by us or other future users of the benchmark is not likely to raise ethical concerns. 

SCOTUS contains personal data of a similar nature. Again, the data is processed and made available by the US Supreme Court, whose proceedings are public. While this ensures compliance with US law, it is very likely that similarly to the ECtHR any processing could be justified by either implied consent or legitimate interest under European law. 

EUR-LEX by contrast is merely a collection of legislation material and therefore not likely to contain personal data, except signatory information (e.g., president of EC). It is openly published by the European Union and processed by the EU’s Publication Office. In addition, since our work qualifies as research, it is privileged pursuant to Art.\ 6 (1) (f) GDPR. 

LEDGAR contains publicly available contract provisions published in the EDGAR database of the US Securities and Exchange Commission (SEC). As far as personal information might be contained, it should equally fall into the public sphere and be covered by research privilege. Our processing does not focus on personal information at all, rather attributing content labels to provisions. 

UNFAIR-ToS contains Terms of Services from business entities such as YouTube, Ebay, Facebook, etc., which makes it unlikely for the data to include personal information. These companies keep user data separate from contractual provisions, so  to the best of our knowledge not contained in this dataset. 

CaseHOLD contains parts of legal decisions from US Court decisions, obtained from the Harvard library case law corpus. All of the decisions were previously published in compliance with US law. In addition, most instances (case snippets) are too short to contain identifiable information. Should such data be contained, their processing would equally be covered either by implicit consent or a public interest exception.
We use all datasets in accordance with copyright terms and under the licenses set forth by their creators. \vspace{2mm}

\noindent\textbf{Limitations \& Potential Harms}\vspace{1mm}

\noindent We have not employed any crowd-workers or annotators for this work. The paper outlines the main limitations with regard to speaker population (English) and generalizability in a dedicated section (Section~\ref{sec:limitations}). As a benchmark paper, our claims naturally match the results of the experiments, which – given the current detail of instructions – should be easily reproduced. We provide several ways of accessing the datasets and running the experiments both with and without Hugging Face infrastructure. 

We do not currently foresee any potential harms for vulnerable or marginalized populations and we do not use, to the best of our knowledge, any identifying characteristics for populations of these kinds.

\bibliography{anthology,acl2020}
\bibliographystyle{acl_natbib}
\appendix

\section{Datasets, Code, and Participation}
\label{sec:codebase}

\noindent\textbf{Where are the datasets?} We provide access to LexGLUE on Hugging Face Datasets \cite{lhoest2021datasets} at \url{https://huggingface.co/datasets/lex_glue}.  For example, to load the SCOTUS dataset, you first simply install the \texttt{datasets} Python library and then make the following call:

{\scriptsize
\begin{verbatim}
___________________________________________________

from datasets import load_dataset 
dataset = load_dataset('lex_glue', task='scotus')

___________________________________________________
\end{verbatim}
}

\noindent\textbf{How do I run experiments?} To make reproducing the results of the already examined models or future models even easier, we release our code on GitHub (\url{https://github.com/coastalcph/lex-glue}). In that repository (in the folder \textsc{/experiments}), there are Python scripts, relying on the Hugging Face Transformers library \cite{wolf-etal-2020-transformers}, to run and evaluate any Transformer-based model (e.g., BERT, RoBERTa, LegalBERT, and their hierarchical variants, as well as, Longformer, and BigBird). We also provide bash scripts to replicate the experiments for each dataset with 5 random seeds, as we did for the reported results.\vspace{2mm}

\begin{table*}[h]
    \centering
    \resizebox{\textwidth}{!}{
    \begin{tabular}{l|c|c|c|c|c|c|c|c}
         \bf Method  & \bf ECtHR (A)* & \bf ECtHR (B)* & \bf SCOTUS* & \bf EUR-LEX & \bf LEDGAR & \bf UNFAIR-ToS  & \bf CaseHOLD\\
         \hline
TFIDF-SVM       & 65.0 & 75.3 & 78.6 & 73.7 & 86.8 & 94.1 & 22.4 \\
BERT            & 71.0 $\pm$ 0.7 & 79.6 $\pm$ 0.5 & 72.7 $\pm$ 0.2 & 77.3 $\pm$ 0.2 & 87.9 $\pm$ 0.1 & \textbf{95.5} $\pm$ 0.0 & 72.8 $\pm$ 0.1\\
RoBERTa         & 70.4 $\pm$ 0.5 & 78.4 $\pm$ 0.7 & 76.9 $\pm$ 0.6 & 77.6 $\pm$ 0.0 & 88.1 $\pm$ 0.1 & 94.8 $\pm$ 0.2 & 74.1 $\pm$ 0.2\\
DeBERTa         & 69.3 $\pm$ 0.7 & 79.0 $\pm$ 0.3 & 76.1 $\pm$ 0.5 & \textbf{77.8} $\pm$ 0.1 & 88.3 $\pm$ 0.2 & \textbf{95.5} $\pm$ 0.1 & 73.8 $\pm$ 0.1\\
\hline
Longformer      & 71.0 $\pm$ 0.3 & \textbf{80.4} $\pm$ 0.9 & 76.9 $\pm$ 0.0 & 77.5 $\pm$ 0.0 & 88.1 $\pm$ 0.2 & 95.1 $\pm$ 0.2 & 73.9 $\pm$ 0.2\\
BigBird         & 71.0 $\pm$ 0.2 & 80.1 $\pm$ 0.5 & 75.9 $\pm$ 0.2 & 77.3 $\pm$ 0.1 & 88.0 $\pm$ 0.1 & 95.2 $\pm$ 0.4 & 73.7 $\pm$ 0.2\\
\hline
Legal-BERT      & 71.9 $\pm$ 0.4 & 79.8 $\pm$ 0.2 & 80.4 $\pm$ 0.3 & 77.6 $\pm$ 0.1 & \textbf{88.5} $\pm$ 0.0 & 95.1 $\pm$ 0.2 & 76.4 $\pm$ 0.3\\
CaseLaw-BERT    & \textbf{72.1} $\pm$ 0.3 & 79.6 $\pm$ 0.0 & \textbf{81.3} $\pm$ 0.6 & 77.2 $\pm$ 0.1 & 88.4 $\pm$ 0.2 & 95.3 $\pm$ 0.4 & \textbf{77.4} $\pm$ 0.2\\
    \end{tabular}
    }
    \vspace{-1mm}
    \caption{Development \microf results for all examined models across all LexGLUE tasks. We report the mean and standard deviations ($\pm$) for the three seeds with the best development scores per model. In starred datasets, we use the hierarchical variant of each model, except for Longformer and BigBird, as discussed in Section~\ref{sec:baselines}.}
    \vspace{-2mm}
    \label{tab:leaderboard_val_micro}
\end{table*}

\begin{table*}[h]
    \centering
    \resizebox{\textwidth}{!}{
    \begin{tabular}{l|c|c|c|c|c|c|c|c}
         \bf Method  & \bf ECtHR (A)* & \bf ECtHR (B)* & \bf SCOTUS* & \bf EUR-LEX & \bf LEDGAR & \bf UNFAIR-ToS  & \bf CaseHOLD\\
         \hline
TFIDF-SVM       & 55.6 & 64.1 & 71.2 & 56.9 & 79.6 & 69.4 & 22.0 \\
BERT            & 65.4 $\pm$ 1.2 & 74.8 $\pm$ 0.6 & 65.9 $\pm$ 0.8 & 62.6 $\pm$ 0.8 & 81.8 $\pm$ 0.1 & 75.8 $\pm$ 1.3 & 72.8 $\pm$ 0.1\\
RoBERTa         & 65.4 $\pm$ 0.2 & 74.2 $\pm$ 1.1 & 69.5 $\pm$ 0.8 & 63.5 $\pm$ 0.4 & 81.9 $\pm$ 0.2 & 74.4 $\pm$ 0.7 & 74.1 $\pm$ 0.2\\
DeBERTa         & 63.5 $\pm$ 0.9 & 74.0 $\pm$ 0.4 & 68.4 $\pm$ 0.8 & 63.6 $\pm$ 0.3 & 82.0 $\pm$ 0.5 & \textbf{77.1} $\pm$ 1.2 & 73.8 $\pm$ 0.1\\
\hline
Longformer      & 65.5 $\pm$ 1.6 & 77.7 $\pm$ 1.0 & 70.4 $\pm$ 0.5 & \textbf{63.8} $\pm$ 0.5 & 82.0 $\pm$ 0.3 & 75.2 $\pm$ 1.2 & 73.9 $\pm$ 0.2\\
BigBird         & 65.8 $\pm$ 1.1 & 74.1 $\pm$ 0.5 & 69.1 $\pm$ 0.2 & 63.0 $\pm$ 0.3 & 81.7 $\pm$ 0.2 & 76.5 $\pm$ 1.8 & 73.7 $\pm$ 0.2\\
\hline
Legal-BERT      & \textbf{68.0} $\pm$ 0.2 & \textbf{76.1} $\pm$ 0.5 & 72.7 $\pm$ 0.2 & 62.0 $\pm$ 0.9 & 82.2 $\pm$ 0.3 & 76.9 $\pm$ 1.3 & 76.4 $\pm$ 0.3\\
CaseLaw-BERT    & 67.1 $\pm$ 0.7 & 74.6 $\pm$ 0.5 & \textbf{74.0} $\pm$ 1.2 & 62.9 $\pm$ 0.3 & \textbf{82.3} $\pm$ 0.3 & 76.5 $\pm$ 0.3 & \textbf{77.4} $\pm$ 0.2\\
    \end{tabular}
    }
    \vspace{-1mm}
    \caption{Development \macrof results for all examined models across all LexGLUE tasks. We report the mean and standard deviation ($\pm$) for the three seeds with the best development scores per model. In starred datasets, we use the hierarchical variant of each model, except for Longformer and BigBird, as discussed in Section~\ref{sec:baselines}.}
    \label{tab:leaderboard_val_macro}
    \vspace{-4mm}
\end{table*}

\section{No labeling as an additional class}
\label{sec:label_zero}

In ECtHR Tasks A \& B and UNFAIR-ToS, there are unlabeled samples. Concretely, in ECtHR Task A, a possible event is \emph{no violation}, i.e., the court ruled that the defendant did not violate any ECHR article. Contrary, \emph{no violation} is not a possible event in the original ECtHR Task B dataset, i.e., at least a single ECHR article is allegedly violated  (considered by the court) in every case; however, there is such a rare scenario after the simplifications we introduced, i.e., some cases were originally labeled only with rare labels that were excluded from our benchmark (Section~\ref{sec:tasks}). In UNFAIR-ToS, the vast majority of sentences are not labeled with any type of \emph{unfairness} (unfair term against users), i.e., most sentences do not raise any questions of possible violations of the European consumer law. 

In multi-label classification, the set of labels per instance is represented as a one-hot vector $Y=[y_1, y_2, \dots, y_L]$, where $y_i = 1$ if the instance is labeled with the $i$-th class, and $y_i = 0$ otherwise. If an instance is not labeled with any class, its $Y$ includes only zeros. During training, binary cross-entropy correctly penalizes such instances, if the predictions ($\hat{Y}=[\hat{y}_1, \hat{y}_2 \dots, \hat{y}_L]$) diverge from zeros. During evaluation, however, the F1-score ($\mathrm{F1}=\frac{\textrm{TP}}{\textrm{TP} + \frac{1}{2}(\textrm{FP} + \textrm{FN})}$) ignores  instances with $Y=\hat{Y}=[0, 0, \dots, 0]$, because it considers only the true positives (\textrm{TP}), false positives (\textrm{FP}), and false negatives (\textrm{FN}), and instances where $Y=\hat{Y}=[0, 0, \dots, 0]$ contribute no \textrm{TP}s, \textrm{FP}s, \textrm{FN}s.
In order to make F1 sensitive to the correct labeling of such examples, during evaluation (not training) we include an additional label ($y_0$ or $\hat{y}_0$) in both targets ($Y$) and predictions ($\hat{Y}$), whose value is 1 (positive) if the original (without $y_0$, $\hat{y}_0$) $Y$ and $\hat{Y}$ are $Y=[0, 0, \dots, 0]$ or $\hat{Y}=[0, 0, \dots, 0]$, respectively, and 0 (negative) otherwise. This is particularly important for proper evaluation, as across three datasets a considerable portion of the examples are unlabeled (11.5\% in ECtHR Task A,  1.6\% in ECtHR Task B, and 95.5\% in UNFAIR-ToS).

\section{Additional Results}
\label{sec:additional}

Tables~\ref{tab:leaderboard_val_micro} and \ref{tab:leaderboard_val_macro} show \emph{development} results for all examined models across datasets. We  report the mean and standard deviations ($\pm$) for the three seeds (among the five used) with the best development scores per model to exclude catastrophic failures, i.e., runs with severely low performance. The standard deviations are relatively low across models and datasets (up to 0.5\% for \microf and up to 1\% for \macrof).
The development results are generally higher compared to the test ones (cf.\ Table~\ref{tab:leaderboard}) in many cases, as one would expect. 

Table~\ref{tab:time} reports training times per dataset and model; both the time per epoch ($T/e$), and the total training time ($T$) across all epochs. All full-attention BERT models, except Longformer and Big-Bird, have comparable times with the exception of DeBERTa that has four separate attention mechanisms. We observe that when the hierarchical variant of these models is deployed, i.e., in ECtHR tasks and SCOTUS, it is approximately twice ($2\times$) as fast compared to Longformer and BigBird.

\begin{table*}[t]
    \centering
    \resizebox{\textwidth}{!}{
    \begin{tabular}{l|cc|cc|cc|cc|cc|cc}
         \multirow{2}{*}{\bf Method}  & \multicolumn{2}{c|}{\bf ECtHR (A)*} & \multicolumn{2}{c|}{\bf ECtHR (B)*} & \multicolumn{2}{c|}{\bf SCOTUS*} & \multicolumn{2}{c|}{\bf EUR-LEX} & \multicolumn{2}{c|}{\bf LEDGAR}  & \multicolumn{2}{c}{\bf CaseHOLD}\\
         & $T$ & $T/e$ & $T$ & $T/e$ & $T$ & $T/e$ & $T$ & $T/e$ & $T$ & $T/e$ & $T$ & $T/e$ \\
         \hline
         BERT & 3h 42m & 28m & 3h 9m & 28m & 1h 24m & 11m & 3h 36m & 19m & 6h 9m & 21m & 4h 24m & 24m \\
         RoBERTa & 4h 11m & 27m & 3h 43m & 27m  & 2h 46m & 17m & 3h 36m & 19m & 6h 22m & 21m & 4h 21m & 24m \\
         DeBERTa & 7h 43m & 46m & 6h 48m & 46m & 3h 42m & 29m & 5h 34m & 36m & 9h 29m & 40m & 6h 42m & 45m \\
         \hline
         Longformer & 6h 47m & 56m & 7h 31m & 56m & 6h 27m & 34m & 11h 10m & 45m & 15h 47m & 50m & 4h 45m & 30m  \\
         BigBird & 8h 41m & 1h 2m & 8h 17m & 1h 2m & 5h 51m & 37m & 3h 57m & 24m & 8h 13m & 27m & 6h 4m & 49m \\
         \hline
         Legal-BERT & 3h 52m & 28m &  3h 2m & 28m & 2h 2m & 17m & 3h 22m & 19m & 5h 23m & 21m & 4h 13m & 23m \\
         CaseLaw-BERT & 3h 2m & 28m & 2h 57m & 28m & 2h 34m & 34m & 3h 40m & 19m & 6h 8m & 21m & 4h 21m & 24m \\
    \end{tabular}
    }
    \caption{Training time in total ($T$) and per epoch ($T/e$) across LexGLUE tasks. In starred datasets, we use the hierarchical variant of each model, except for Longformer and BigBird, as described in Section~\ref{sec:baselines}.}
    \label{tab:time}
    \vspace{-2mm}
\end{table*}

\begin{table*}[t]
    \centering
    \resizebox{\textwidth}{!}{
    \begin{tabular}{l|cc|cc|cc|cc|cc|cc|c|c}
         \multirow{2}{*}{\bf Method}  & \multicolumn{2}{c|}{\bf ECtHR (A)*} & \multicolumn{2}{c|}{\bf ECtHR (B)*} & \multicolumn{2}{c|}{\bf SCOTUS*} & \multicolumn{2}{c|}{\bf EUR-LEX} & \multicolumn{2}{c|}{\bf LEDGAR}  & \multicolumn{2}{c|}{\bf UNFAIR-ToS} & \bf CaseHOLD\\
         & \microf & \macrof & \microf & \macrof & \microf & \macrof & \microf & \macrof & \microf & \macrof & \microf & \macrof & \microf / \macrof \\
         \hline
\multicolumn{14}{c}{\bf Results on Development Set} \\
\hline
RoBERTa (B)     &  70.6 &  65.7 & 79.3 &  75.8 & 77.5 & \bf 64.1 & 77.6 &  70.4 & 88.0 &  82.1 & 94.6 &  75.2 & 74.3 \\
RoBERTa (L)     &  \bf 72.7 &  \bf 69.3 &  \bf 81.1 &  \bf 77.0 & 74.6 &  56.9 & 78.0 &  74.5 & 88.5 & \bf 82.8 & \bf 95.8 &  \bf 80.3 & 76.8\\
\hline
Legal-BERT      & 72.5 &  68.2 & 79.7 &  76.8 & \bf 77.6 &  63.3 & 80.8 &  72.9 & 88.5 &  82.6 & 95.3 &  78.2 & 76.6 \\
CaseLaw-BERT    & 71.8 &  67.7 & 79.5 &  74.9 & 77.3 &  63.1 & \bf 82.1 &  \bf 75.6 & \bf 88.7 &  82.7 & 95.7 &  76.9 & \bf 77.7  \\
\hline
\multicolumn{14}{c}{\bf Results on Test Set} \\
\hline
RoBERTa (B)     & 69.2 & 59.0 & 77.3 & 68.9 & 71.6 & 62.0 & 71.9 & 57.9 & 87.9 & 82.3 & 95.2 & 79.2 & 71.4 \\
RoBERTa (L)     & \bf 73.8 & \bf 67.6 & 79.8 & 71.6 & 75.5 & 66.3 & \bf 72.5 & \bf 58.1 & \bf 88.6 & \bf 83.6 & 95.8 & 81.6 & 74.4 \\
\hline
Legal-BERT      & 70.0 & 64.0 & \bf 80.4 & \bf 74.7 & 76.4 & \bf 66.5 & 72.1 & 57.4 & 88.2 & 83.0 & \bf 96.0 & \bf 83.0 & 75.3 \\
CaseLaw-BERT    & 69.8 & 62.9 & 78.8 & 70.3 & 76.6 & 65.9 & 70.7 & 56.6 & 88.3 & 83.0 & \bf 96.0 & 82.3 & \bf 75.4 \\
    \end{tabular}
    }
    \vspace{-1mm}
    \caption{Development and test results across LexGLUE tasks. In starred datasets, we use the hierarchical variant of each model, discussed in Section~\ref{sec:baselines}. (B) and (L) denote the base and large version of RoBERTa, respectively.}
    \label{tab:roberta_large}
    \vspace{-4mm}
\end{table*}

\begin{figure}[h]
    \centering
    \resizebox{\columnwidth}{!}{
    \includegraphics{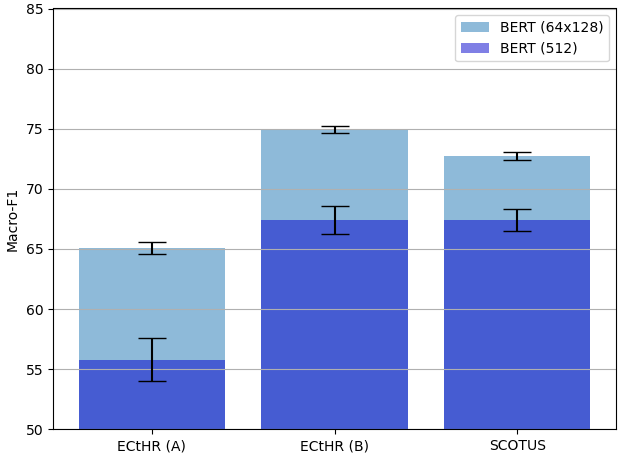}
    }
    \caption{Development \macrof scores of standard BERT (up to 512 tokens) and its hierarchical variant (Section~\ref{sec:pretrainedModels}, 64$\times$128 tokens) in ECtHR (Task A, B) and SCOTUS, i.e., the datasets with long documents. Light blue denotes the average score across 5 runs for the hierarchical variant (used in Table~\ref{tab:leaderboard} for these datasets), while dark blue corresponds to standard BERT (not used in in Table~\ref{tab:leaderboard} for these datasets). The error bars show the standard error.
    }
    \vspace{-2mm}
    \label{fig:bert_base_truncated}
    \vspace{-4mm}
\end{figure}

\section{Use of 512-token BERT models}
\label{sec:bert_truncated}
In Figure~\ref{fig:bert_base_truncated}, we show results for the standard BERT model of \citet{devlin-etal-2019-bert}, which can process up to 512 tokens, compared to its hierarchical variant (Section~\ref{sec:pretrainedModels}), which can process up to 64$\times$128 tokens. We observe that across all datasets that contain long documents (ECtHR A \& B, SCOTUS, cf.\ Fig.~\ref{fig:lengths}(a)), the hierarchical variant clearly outperforms the standard model fed with truncated documents (ECtHR A: +10.2\% p.p., ECtHR B: 7.5\% p.p., SCOTUS: 4.9\% p.p.). Compared to the ECtHR tasks, the gains are lower in SCOTUS, a topic classification task where long-range reasoning is not needed; by contrast, for ECtHR multiple distant facts need to be combined. Based on these results, we conclude that using severely truncated documents is not a plausible option for LexGLUE, and other directions for processing long documents should be considered in the future, ideally fully pre-trained hierarchical models, contrary to our semi-pre-trained hierarchical models (Section~\ref{sec:vision}).

\section{Use of Roberta Large}
\label{sec:roberta-large}

We additionally evaluate RoBERTa-large, i.e., 24 Transformer blocks, 1024 hidden units, and 18 attention heads, to better understand the dynamics between domain specificity and model size. In this case, we use the AdamW optimizer with a 1e-5 maximum learning rate,  warm-up ratio of 0.1, and a weight decay rate of 0.06, and we use a similar mini-batch size of 8 examples.\footnote{Large models tend to be very sensitive to parameter updates, especially in the initial training steps; hence a smaller learning rate and warm up steps are very crucial.}

Table~\ref{tab:roberta_large} reports the development and test results using the seed (run) with the best development scores. We observe that using the large version of RoBERTa, dubbed RoBERTa (L), with more than 2$\times$ parameters (355M),  leads to substantial performance improvements  compared  to  the  base  version  of RoBERTa, dubbed RoBERTa (B), across all tasks. 
The results are comparable, or better in some cases, compared to the legal-oriented language models (Legal-BERT, CaseLaw-BERT). 

Considering that the two legal-oriented models are much smaller and have been pre-trained with ($5\!-\!10\times$) less data (Section~\ref{tab:examined_models}), we have a strong indication for expected performance gains by pre-training larger legal-oriented models using larger legal corpora (Section~\ref{sec:vision}).

\section{Other Tasks and Datasets Considered}

We considered including the Contract Understanding Atticus Dataset (CUAD) \cite{Hendrycks2021CUAD}, an expertly curated dataset that comprises 510 contracts annotated with 41 valuable contractual insights (e.g., agreement date, parties, governing law). The task is formulated as a SQUAD-like question answering task, where given a \emph{question} (the name of an insight) and a \emph{paragraph} from the contract, the model has to identify the answer span in the paragraph.\footnote{The question mostly resembles a \emph{prompt}, rather than a natural question, as there is a closed set of 41 alternatives.} 
The original dataset follows the SQUAD v2.0 setting, including unanswerable questions.
Following SQUAD v1.1 \cite{rajpurkar-etal-2016-squad}, we simplified the task by removing all unanswerable pairs (question, paragraph), which are the majority in the original dataset. We also excluded pairs whose answers exceeded 128 full words to alleviate the imbalance between short and long answers. We then re-split the dataset chronologically into training (5.2k, 1994--2019), development (572, 2019--2020), and test (604, 2020) sets.

Following \citet{devlin-etal-2019-bert}, and similarly to \citet{Hendrycks2021CUAD}, for each training (or test) instance, we consider pairs that consist of a question and a paragraph, separated by the special delimiter token \sep. The top-level representations $[h_1, \dots, h_N]$ of the tokens of the paragraph are fed into a linear layer to obtain two logits per token (for the token being the start or end of the answer span), which are then  passed through a softmax activation (separately for start and end) to obtain probability distributions. The tokens with the highest start and end probabilities are selected as boundaries of the answer span. We evaluated performance with token-level F1 score, similarly to SQUAD.

We trained all the models of Table~\ref{tab:examined_models}, which scored approx.\ 10-20\% in token-level F1, with Legal-BERT performing slightly better than the rest (+5\% F1).\footnote{F1 is one of the two official SQUAD measures. In the second one, Exact Answer Accuracy, all models scored 0\%.}  In the paper that introduced CUAD  \cite{Hendrycks2021CUAD}, several other measures (Precision@ N\% Recall, AUPR, Jaccard similarity) are used to more leniently estimate a  model's ability to approximately locate answers in context paragraphs. Through careful manual inspection of the dataset, we noticed the following points that seem to require more careful consideration.
\begin{itemize}[leftmargin=8pt]
    \item Contractual insights (categories, shown in italics below) include both entity-level (short) answers (e.g., ``SERVICE AGREEMENT'' for \emph{Document Name}, and ``Imprimis Pharmaceuticals, Inc.'' for \emph{Parties}) and paragraph-level (long) answers (e.g., ``If any of the conditions  specified in Section 8 shall not have been  fulfilled when and as required by this  Agreement,  or by the Closing Date,  or waived in writing  by Capital  Resources,  this  Agreement  and all of Capital Resources  obligations hereunder may be canceled [...] except as otherwise  provided in Sections 2, 7, 9 and 10 hereof.'' for \emph{Termination for Convenience}). These two different types of answers (short and paragraph-long) seem to require different models and different evaluation measures, unlike how they are treated in the original CUAD paper.
    
    \item Some contractual insights (categories), e.g., \emph{Parties}, have been annotated with both short (e.g., ``Imprimis Pharmaceuticals, Inc.'') and long (e.g., ``together, Blackwell and Munksgaard shall be referred to as `the Publishers'.'') answers. 
    Annotations of this kind introduce noise during both training and evaluation. For example, it becomes unclear when a short (finer/strict) or a long (loose) annotation should be taken to be the correct one. 
    
    \item Annotations may include indirect mentions, e.g., `Franchisee', `Service Provider' for \emph{Parties},  instead of the actual entities (the company name). 
    
    \item Annotations may include semi-redacted text (e.g., ``\underline{\hspace{1cm}}, 1996'' for \emph{Agreement Date}), or even fully redacted text (e.g., ``\underline{\hspace{2cm}}'' for \emph{Parties}). This practice may be necessary to hide sensitive information, but for the purposes of a benchmark dataset such cases could have been excluded.
\end{itemize}

\noindent 
The points above, which seem to require revisiting the annotations of CUAD, and the very low F1 scores of all models led us to exclude CUAD from LexGLUE. We also note that there is related work covering similar topics, such as Contract Element Extraction \cite{Chalkidis2017Jurix}, Contractual Obligation Extraction \cite{chalkidis-etal-2018-obligation}, and Contractual Provision Classification \cite{tuggener-etal-2020-ledgar}, where models perform much better (in terms of accuracy), relying on simpler (separate) more carefully designed tasks and much bigger datasets. Thus we believe that the points mentioned above, which blur the task definition of CUAD and introduce noise, and the limited (compared to larger datasets) number of annotations strongly affect the performance of the models on CUAD, underestimating their true potential.

\medskip We also initially considered some very interesting legal Information Retrieval (IR) datasets \cite{locke-2018,chalkidis-etal-2021-regulatory} that aim to examine crucial real-life tasks (relevant case law retrieval, regulatory compliance). However, we decided to exclude them from the first version of LexGLUE, because they rely on processing multiple long documents and require more task-specific neural network architectures (e.g., siamese networks), and different evaluation measures. Hence, they would make LexGLUE more complex and a less attractive entry point for newcomers to legal NLP. We plan, however, to include more demanding tasks in future LexGLUE versions, as the legal NLP community will be growing.

\begin{table*}[t]
    \centering
    \resizebox{\textwidth}{!}{
    \begin{tabular}{l|p{20cm}|l}
        \bf Dataset & \bf Input(s) & \bf Output(s) / Label(s) \\
        \hline
         ECtHR & \specialcelll{\textbf{Text:} 12. In 1987 the applicant association published a book entitled Euskadi at war. There were four versions – Basque, English, Spanish and French – and the book was distributed in numerous countries, including France and Spain. According to the applicant association, this was a collective work containing contributions from a number of academics with specialist knowledge of the Basque Country and giving an account of the historical, cultural, linguistic and socio-political aspects of the Basque cause. It ended with a political article entitled “Euskadi at war, a promise of peace” by the Basque national liberation movement.\\ 13. The book was published in the second quarter of 1987. On 29 April 1988 a ministerial order was issued by the French Ministry of the Interior under section 14 of the Law of 29 July 1881, as amended by the decree of 6 May 1939, banning the circulation, distribution and sale of the book in France in any of its four versions on the ground that “the circulation in France of this book, which promotes separatism and vindicates recourse to violence, is likely to constitute a threat to public order”. On 6 May 1988, pursuant to the aforementioned order, the département director of the airport and border police refused to allow over two thousand copies of the book to be brought into France. \emph{[...]}} & \specialcell{3 (\emph{Right to a fair trial}) \\ 6 (\emph{Freedom of expression})}\\
         \hline
         SCOTUS & \specialcelll{\textbf{Text:} 329 U.S. 29 67 S.Ct. 1 91 L.Ed. 22 \\
         CHAMPLIN REFINING CO v. UNITED STATES et al. No. 21. Argued Oct. 18, 21, 1946. Decided Nov. 18, 1946.\\
         Appeal from the District Court of the United States for the Western District of Oklahoma. Messrs. Dan Moody, of Austin, Tex., and Harry O. Glasser, of Enid, Okla., for appellant. Mr. Edward Dumbauld, of Washington, D.C., for appellees. \\ Mr. Justice JACKSON delivered the opinion of the Court.\\ 1 The Interstate Commerce Commission, acting under § 19a of the Interstate Commerce Act,1 ordered the appellant to furnish certain inventories, schedules, maps and charts of its pipe line property.\\ 2 Champlin's objections that the Act does not authorize the order, or if it be construed to do so is unconstitutional, were overruled by the Commission and again by the District Court which dismissed the company's suit for an injunction.3 These questions of law are brought here by appeal. \emph{[...]} }& 7 (\emph{Economic Activity}) \\
         \hline
         \specialcell{EUR-LEX} & \specialcelll{\textbf{Text:} Commission Regulation (EC) No 1156/2001 of 13 June 2001 fixing the export refunds on white sugar and raw sugar exported in its unaltered state \\ THE COMMISSION OF THE EUROPEAN COMMUNITIES \\ Having regard to the Treaty establishing the European Community, Having regard to Council Regulation (EC) No 2038/1999 of 13 September 1999 on the common organisation of the markets in the sugar sector(1), as amended by Commission Regulation (EC) No 1527/2000(2), and in particular point (a) of the second subparagraph of Article 18(5) thereof, \\ Whereas: (1) Article 18 of Regulation (EC) No 2038/1999 provides that the difference between quotations or prices on the world market for the products listed in Article 1(1)(a) of that Regulation and prices for those products within the Community may be covered by an export refund. (2) Regulation (EC) No 2038/1999 provides that when refunds on white and raw sugar, undenatured and exported in its unaltered state, are being fixed account must be taken of the situation on the Community and world markets in sugar and in particular of the price and cost factors \emph{[...]}} & \specialcell{28 (\emph{Trade Policy}),\\  93  (\emph{Beverages and Sugar}),\\  94  (\emph{Foodstuff})} \\
         \hline 
         LEDGAR & \textbf{Text:} The validity or unenforceability of any provision or provisions of this Agreement shall not affect the validity or enforceability of any other provision hereof, which will remain in full force and effect.  Should a court or other body of competent jurisdiction determine that any provision of this Agreement is excessive in scope or otherwise illegal, invalid, void or unenforceable, such provision shall be adjusted rather than voided, if possible, so that it is enforceable to the maximum extent possible. & 79 (Severability) \\
         \hline
         UNFAIR-ToS & \textbf{Text:} By creating a tinder account or by using the tinder imessage app (``tinder stacks''), whether through a mobile device , mobile application or computer (collectively, the ``service'') you agree to be bound by (i) these terms of use, (ii) our privacy policy and safety tips, each of which is incorporated by reference into this agreement, and (ii ) any terms disclosed and agreed to by you if you purchase additional features, products or services we offer on the service (collectively, this ``agreement''). & 4 (\emph{Contract by Using}) \\
         \hline
          CaseHOLD & \specialcelll{\textbf{Context:} Drapeau’s cohorts, the cohort would be a “victim” of making the bomb. Further, firebombs are inherently dangerous. There is no peaceful purpose for making a bomb. Felony offenses that involve explosives qualify as “violent crimes” for purposes of enhancing the sentences of career offenders. See 18 U.S.C. § 924(e)(2)(B)(ii) (defining a “violent felony” as: “any crime punishable by imprisonment for a term exceeding one year ... that ... involves use of explosives”). Courts have found possession of a'bomb to be a crime of violence based on the lack of a nonviolent purpose for a bomb and the fact that, by its very nature, there is a substantial risk that the bomb would be used against the person or property of another. See United States v. Newman, 125 F.3d 863 (10th Cir.1997) (unpublished) ([HOLDING]); United States v. Dodge, 846 F.Supp. 181 \\ \textbf{Choices (Holdings):} \\ 
          (A) "holding that possession of a pipe bomb is a crime of violence for purposes of 18 usc 3142f1", \\
          (B) "holding that bank robbery by force and violence or intimidation under 18 usc 2113a is a crime of violence", \\
          (C) "holding that sexual assault of a child qualified as crime of violence under 18 usc 16", \\
          (D) "holding for the purposes of 18 usc 924e that being a felon in possession of a firearm is not a violent felony as defined in 18 usc 924e2b", \\
          (E) "holding that a court must only look to the statutory definition not the underlying circumstances of the crime to determine whether a given offense is by its nature a crime of violence for purposes of 18 usc 16"} & \specialcell{0 (\emph{Choice A})} \\ 
         
    \end{tabular}
    }
    \caption{Training examples (pairs of inputs, outputs) for LeXGLUE datasets and tasks.}
    \label{tab:dataset_examples}
\end{table*}

\section{Dataset Examples}
\label{sec:dataset_examples}

In Table~\ref{tab:dataset_examples}, we present training examples, i.e., pairs of input(s), output(s), for LeXGLUE datasets and tasks. More examples can be inspected using the dataset preview functionality provided in the online dataset card of Hugging Face.\footnote{\url{https://huggingface.co/datasets/lex_glue}} \clearpage

\section{Updated version V4 (09/11/2022)}

We updated the results for the TFIDF-SVM method in Tables~\ref{tab:leaderboard}, \ref{tab:leaderboard_val_micro} and~\ref{tab:leaderboard_val_macro}. There was a bug in our code base,\footnote{\url{https://github.com/coastalcph/lex-glue}} where TFIDF-SVM grid search function was re-fitting (re-training) the model given both the training and development sets. This was inconsistent with our general practices leading to overestimated development and test scores. The new corrected results make clear that TFIDF-SVM models are significantly outperformed in most cases, even on the SCOTUS dataset. Thanks to Yu-Chen and Daniel Gonzalez for pointing out this issue.

\end{document}